\documentclass[sigconf]{acmart} 
\AtBeginDocument{%
  }

\acmConference[ ]{ }{ }

\renewcommand\footnotetextcopyrightpermission[1]{} 
\newcommand{\ie}{\textit{i}.\textit{e}.}

\usepackage{pifont}
\usepackage{colortbl}
\usepackage{arydshln}
\usepackage{multirow}
\usepackage{multicol}
\usepackage{mathrsfs}
\usepackage{bm}
\begin{document}
\title{Diffusion-Guided Knowledge Distillation for Weakly-Supervised Low-Light Semantic Segmentation}
\author{Chunyan Wang}
\affiliation{
  \institution{Nanjing University of Science and Technology}
  \city{Nanjing}
   \country{China}
}
\email{carrie\_yan@njust.edu.cn}

\author{Dong Zhang}
\affiliation{
  \institution{The Hong Kong University of Science and Technology}
  \city{Hong Kong}
   \country{China}
}
\email{dongz@ust.hk}

\author{Jinhui Tang}
\authornote{Corresponding author.}
\affiliation{
  \institution{Nanjing Forestry University}
  \city{Nanjing}
  \country{China}
}
\email{tangjh@njfu.edu.cn}

\begin{abstract}
Weakly-supervised semantic segmentation aims to assign category labels to each pixel using weak annotations, significantly reducing manual annotation costs. Although existing methods have achieved remarkable progress in well-lit scenarios, their performance significantly degrades in low-light environments due to two fundamental limitations: severe image quality degradation (e.g., low contrast, noise, and color distortion) and the inherent constraints of weak supervision. These factors collectively lead to unreliable class activation maps and semantically ambiguous pseudo-labels, ultimately compromising the model's ability to learn discriminative feature representations. To address these problems, we propose Diffusion-Guided Knowledge Distillation for Weakly-Supervised Low-light Semantic Segmentation (DGKD-WLSS), a novel framework that synergistically combines Diffusion-Guided Knowledge Distillation (DGKD) with Depth-Guided Feature Fusion (DGF2). DGKD aligns normal-light and low-light features via diffusion-based denoising and knowledge distillation, while DGF2 integrates depth maps as illumination-invariant geometric priors to enhance structural feature learning. Extensive experiments demonstrate the effectiveness of DGKD-WLSS, which achieves state-of-the-art performance in weakly supervised semantic segmentation tasks under low-light conditions. The source codes have been released at:~\href{https://github.com/ChunyanWang1/DGKD-WLSS}{DGKD-WLSS}.
\end{abstract}

\maketitle
\section{Introduction}
\begin{figure}[tb]
\includegraphics[width=.45 \textwidth]{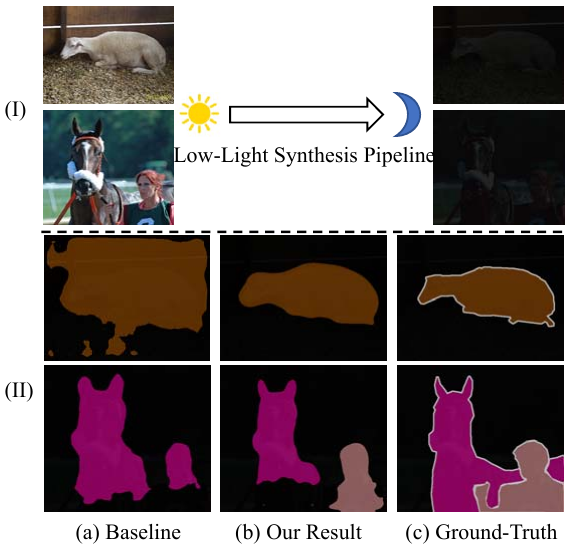}
\vspace{-2mm}
\caption{The challenges of weakly supervised low-light semantic segmentation: (I) dataset deficiency and (II) low contrast in low-light images lead to the problems of low-confident CAMs (e.g. over-activation) and semantic confusion(e.g. misidentifying the ``horse" as a ``person"). To address the first challenge, we synthesize realistic low-light images from the PASCAL VOC 2012~\cite{everingham2010pascal} dataset using a low-light synthesis pipeline~\cite{brooks2019unprocessing,cui2021multitask}. For the second challenge, our DGKD-WSSS introduces DGKD and DGF2 modules to resolve these issues, generating more accurate segmentation results.
}
\label{motivation}
\vspace{-4mm}
\end{figure}

Weakly Supervised Semantic Segmentation (WSSS) is a fundamental task in computer vision that aims to assign object category labels to each pixel in an image using weak supervision annotations, thereby significantly reducing manual annotation costs. Common weak supervision forms include bounding boxes~\cite{dai2015boxsup,kulharia2020box2seg}, scribbles~\cite{lin2016scribblesup,zhang2024scribble}, points~\cite{bearman2016s,zhao2024point} and image-level labels~\cite{ahn2018learning,zhang2020causal,ru2023token}. Among these, image-level annotations are the most widely adopted due to their ease of acquisition. Although deep learning-based WSSS methods have achieved remarkable progress on well-illuminated datasets, low-light WSSS with image-level labels remains largely unexplored. This research gap stems from two primary factors: 1) Dataset deficiency: Existing mainstream benchmarks (e.g., PASCAL VOC 2012~\cite{everingham2010pascal}, COCO 2014~\cite{lin2014microsoft}) predominantly focus on normal illumination conditions, lacking specialized datasets with pixel-level annotations for low-light environments. Although several nighttime datasets have been developed (e.g., Dark Zurich~\cite{sakaridis2019guided}, ACDC\_Night~\cite{sakaridis2021acdc}, NightCity~\cite{tan2021night}) to advance supervised nighttime segmentation, these remain constrained to driving scenarios and require pixel-level supervision for training; 2) Feature learning: Conventional image-level WSSS approaches rely on Class Activation Maps (CAMs) for object localization. However, low-light images suffer from degradations like low contrast, noise, and color distortion, leading to two critical issues: (1) \textbf{Semantic confusion}: illumination-affected pseudo-masks may mislead classifiers to focus on non-target areas or misidentify objects during iterative optimization. For instance, in Fig.~\ref{motivation}(\uppercase\expandafter{\romannumeral2}), the model confuses a "person" with a "horse", impairing its ability to learn semantically consistent features. (2) \textbf{Low-confidence CAMs}: The lack of clear structural features in dark images results in unreliable CAMs. As shown in Fig.~\ref{motivation}(\uppercase\expandafter{\romannumeral2}), the low contrast between the ``sheep" and background prevents the model from distinguishing its boundaries, causing activation spillover into background regions.

\begin{figure}[tb]
\includegraphics[width=.48 \textwidth]{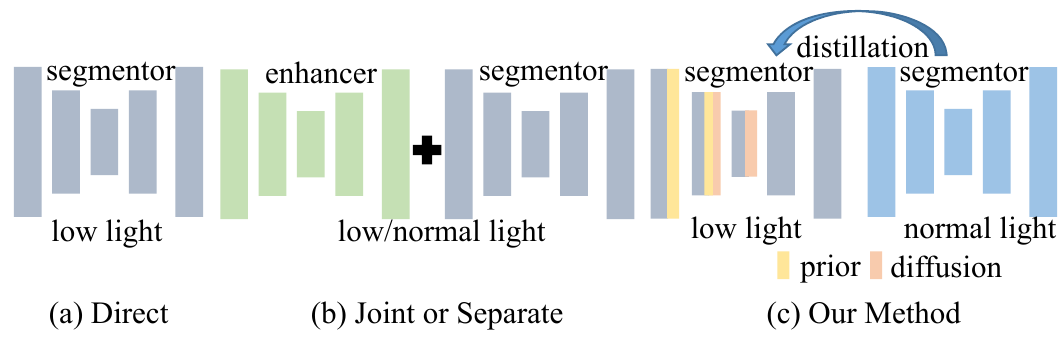}
\caption{Comparisons of frameworks for weakly supervised semantic segmentation under low-light conditions. Different to low-light direct (a) and enhancement (b) methods, we introduce the diffusion model for knowledge distillation to denoise low-light features, augmented by depth prior for structural learning, enabling better cross-illumination knowledge transfer. We offer a novel solution for low-light WSSS.}
\label{fig1}
\end{figure}

Therefore, to achieve weakly supervised semantic segmentation in low-light conditions, we must address two key challenges. The first challenge is the lack of naturally captured low-light datasets with image-level annotations. The second challenge lies in how to extract well-structured and semantically consistent features from low-light images using only image-level supervision? To overcome the dataset deficiency, we adopt a low-light synthesis pipeline~\cite{brooks2019unprocessing,cui2021multitask} (shown in (\uppercase\expandafter{\romannumeral1}) of Fig.~\ref{motivation})to generate realistic low-light datasets from existing well-lit natural RGB image datasets (e.g., PASCAL VOC 2012 ~\cite{everingham2010pascal}). This approach enables weakly supervised semantic segmentation training under low-light conditions. For the second challenge, a common approach is to simulate fully supervised low-light segmentation by first enhancing the dark images using low-light enhancement methods~\cite{wei2018deep,jiang2021enlightengan,lamba2021restoring,zhang2025wakeup}, then training a segmentation model on the enhanced outputs. However, most existing methods follow this two-stage pipeline, which may introduce artifacts or suboptimal segmentation performance. Given the remarkable progress of weakly supervised semantic segmentation models trained on normal-light images, we explore whether such pre-trained models can guide low-light images in learning semantically consistent features. We propose a novel approach: leveraging knowledge from models trained on well-lit images to segment low-light images. As shown in Fig.~\ref{fig1}, our framework differs from traditional low-light enhancement methods. Instead, we employ knowledge distillation~\cite{hinton2014distilling,zhang2024cross} to transfer semantic knowledge from normal-light features to low-light ones, thereby learning semantically aligned representations. However, distillation methods~\cite{hinton2014distilling,gou2021knowledge,huang2022knowledge} can improve performance, while they struggle to align features with large distributional discrepancies caused by illumination variations. Inspired by DiffKD~\cite{huang2024knowledge}, we assume that low-light features are essentially noisy variants of normal-light features due to illumination degradation. Thus, we propose to integrate a diffusion model into the distillation process to systematically denoise low-light features, generating clean features that closely resemble those from well-lit images. This enables robust cross-illumination knowledge transfer, offering a novel solution for low-light WSSS. To further improve structural feature learning, we incorporate Depth Anything~\cite{yang2024depth} to provide depth maps as illumination-invariant geometric priors. Depth information serves as an additional modality, helping the model capture precise object structures despite lighting variations, thereby enhancing robustness in low-light conditions.

Specifically, we propose Diffusion-guided Knowledge Distillation for Weakly-Supervised Low-light Semantic Segmentation (DGKD-WLSS), which comprises two core modules: Diffusion-Guided Knowledge Distillation (DGKD) and Depth-Guided Feature Fusion (DGF2). The DGKD module employs a diffusion model to denoise low-light features while distilling knowledge from normal-light features, enabling effective cross-illumination feature alignment. Meanwhile, the DGF2 module integrates visual priors (depth maps) with low-light features to learn structured representations, thereby enhancing semantic perception and assisting the distillation process. To the best of our knowledge, this work presents the first systematic exploration of weakly supervised segmentation under low-light conditions, offering novel insights for semantic understanding in challenging illumination scenarios. Experimental results demonstrate our method's superior performance on both synthetic low-light PASCAL VOC 2012 and the real-world LIS~\cite{chen2023instance} dataset. The main contributions of this paper can be concluded as:

(1) We employ a synthetically darkened PASCAL VOC 2012 dataset generated through a low-light synthesis pipeline~\cite{brooks2019unprocessing,cui2021multitask} for training, while evaluating on real-world low-light datasets, thereby validating the efficacy of our method for weakly supervised low-light segmentation tasks.

(2) We design a DGKD module, which is utilized to align normal-light (teacher) and low-light (student) features through diffusion-based denoising. It can effectively remove noise from low-light features while enabling the student network to learn effective representations from the teacher.

(3) We design a DGF2 module, which fuses illumination-invariant depth priors with low-light features to learn comprehensive structural representations. It can enhance boundary awareness and robustness in challenging lighting conditions.

\section{Related Work}
\subsection{Semantic Segmentation in the Low-Light}
To adopt semantic segmentation~\cite{zhang2020feature,chen2023boosting,zhang2025generalized} for low-light scenes, a straightforward solution is utilizing the low-light enhancement methods~\cite{wei2018deep,jiang2021enlightengan,lamba2021restoring,ji2023multi,zhang2025wakeup} as a pre-processing step. However, these methods require independent training before integration into semantic segmentation, which adds training constraints. 
To simplify training, DIAL-Filters~\cite{liu2023improving} employs a lightweight network jointly trained with the segmentation model to adaptively enhance low-light inputs, using only a segmentation loss. Meanwhile, ESSNLL~\cite{mu2024end} introduces a Dual Closed-loop Bipartite Matching algorithm to resolve conflicts between enhancement and segmentation losses, enabling joint optimization.
In addition, early approaches employed domain adaptation~\cite{dai2018dark,sakaridis2019guided,sakaridis2020map,xu2023mada} to transfer semantic knowledge from well-illuminated to low-light scenes. However, with the introduction of large-scale datasets like NightCity~\cite{tan2021night}, research shifted toward fully-supervised learning. For instance, NightLab~\cite{deng2022nightlab} improves segmentation by classifying objects into simple/difficult categories and prioritizing challenging regions through a hardness detection mechanism.
Furthermore, DTP~\cite{wei2023disentangle} disentangles illumination and content features to enable illumination-invariant segmentation. However, most current works focus on driving scenarios and require pixel-level annotations. In contrast, we propose transferring knowledge from well-illuminated weakly-supervised models to low-light images via diffusion-based feature denoising, enhancing model generalization.

\subsection{Low-Light Synthesis}
The goal of low-light image synthesis is to enhance or generate images captured under poor lighting conditions, improving their visual quality or facilitating downstream tasks (e.g., object detection~\cite{wang2024yolov10,zhao2024detrs} or segmentation~\cite{strudel2021segmenter,zhou2024cross}). However, most low-light enhancement methods~\cite{wei2018deep,chen2018learning,liu2021retinex,cai2023retinexformer} typically require paired low-light/normal-light images for training, which are challenging to acquire in real-world scenarios. To address this limitation, researchers have explored various approaches~\cite{wei2018deep,wang2018gladnet,cui2021multitask,wei2021physics,punnappurath2022day} for synthesizing low-light images from normal-light counterparts. Among these, RetinexNet~\cite{wei2018deep} developed a method utilizing normal-light RAW images from the RAISE~\cite{dang2015raise} dataset, where the histogram of the Y channel in YCbCr color space was adjusted to match low-light characteristics from public datasets, subsequently generating synthetic low-light images. Similarly, GLADNet~\cite{wang2018gladnet} implemented synthesis approach using RAW images by manipulating exposure, vibrance, and contrast parameters. Drawing inspiration from recent advancements~\cite{cui2021multitask,wei2021physics} in this field, which have significantly improved low-light image synthesis by incorporating noise into their frameworks, our approach focuses on synthesizing low-light RGB images directly from normal-light RGB images of natural scenes, incorporating quantisation noise to enhance the realism of the synthesized low-light conditions. 

\subsection{Knowledge Distillation}
Knowledge distillation (KD) is an effective method for transferring knowledge from a large, complex model (teacher) to a smaller, efficient model (student). In low-light image enhancement task, KD improves performance by addressing challenges like noise, low contrast, and ambiguous boundaries of low-light images. Ko et al.~\cite{ko2021learning} presented a lightweight enhancement network trained through KD, using pseudo well-exposed images for real-world low-light enhancement. Park et al.~\cite{park2022dual} extended the Retinex framework with a dual-teacher distillation model, introducing an attention-based mechanism for feature extraction. It can improve low-light image brightness and segmentation accuracy. Jeong et al.~\cite{jeong2023low} proposed a model that distills knowledge from near-infrared (NIR) to RGB conversion networks. This enhances low-light images by preserving details and reducing noise. Different to above methods, we propose to utilize the KD to transfer the valuable semantic information obtained from the well-illuminated features to low-light ones with the help of diffusion models.

\subsection{Diffusion Models in Low-Light Scenes}
Recent studies have demonstrated that diffusion models achieve promising performance in low-light image enhancement tasks~\cite{jiang2023low,zhou2023pyramid,hou2024global}, which directly benefits downstream applications such as semantic segmentation and object detection in low-light environments. Most existing approaches adopt a two-stage framework, where diffusion models are first employed to enhance low-light images before passing them to segmentation or detection networks for further processing. For instance, GSAD~\cite{hou2024global} adopted a structure-aware diffusion process, incorporating global curvature regularization to stabilize the diffusion trajectory, which can reduce noise artifacts in low-light images.
WCDM~\cite{jiang2023low} was proposed to combine wavelet transformation with diffusion processes to retain high-frequency details while suppressing noise, which can efficiently enhance the low-light images.
PyDiff~\cite{zhou2023pyramid} introduced a hierarchical pyramid diffusion approach where low-light images are processed progressively from low to high resolution, mitigating the color shifts and preserve image details. 
These methods can significantly improve the following segmentation or detection accuracy in low-light environments. However, there are few studies that directly adopt the diffusion models on the weakly-supervised low-light segmentation network, especially considering the low-light segmentation as a denosing problem. In this paper, we proposed to utilize the diffusion model to recover the knowledge hidden in the low-light features during the distillation process.

\section{Method}
\begin{figure*}[tb]
\includegraphics[width=.98 \textwidth]{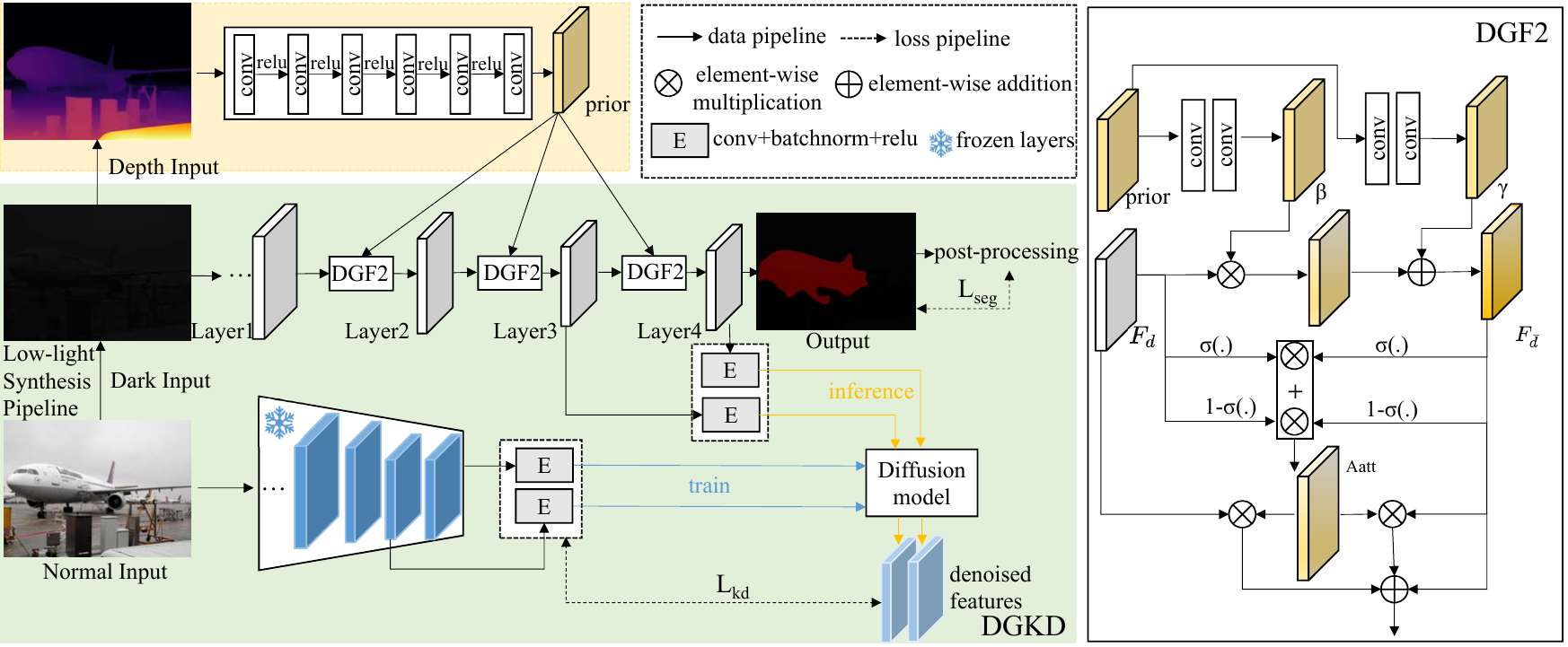}
\vspace{-2mm}
\caption{Our proposed DGKD-WLSS framework addresses the challenges of semantic confusion and low-confidence CAMs in weakly supervised low-light semantic segmentation. The framework comprises two key modules: (1) the Diffusion-Guided Knowledge Distillation (DGKD), which transfers semantic knowledge from normal-light to low-light features to ensure semantically consistent representations; and (2) the Depth-Guided Feature Fusion (DGF2), which leverages depth maps as auxiliary priors to improve structural feature representation, thereby generating reliable CAMs and refining object boundaries. Here, the terms ``train" and ``inference" specifically refer to the diffusion model rather than DGKD-WLSS. Notably, the diffusion model is not utilized during the inference phase of our framework.  }
\label{fig2}
\vspace{-2mm}
\end{figure*}

In this section, we formulate our proposed DGKD-WLSS method. First, we give the preliminaries about the diffusion model. Then, we introduce the framework of distilling the semantic knowledge of normal-light features to low-light ones by denoising the low-light ones with a diffusion model (i.e., DGKD module). Finally, to further learn more structural feature representations, we introduce depth maps as visual prior knowledge to provide the geometric information and then fuse it with low-light features (i.e., DGF2 module). The overall architecture of our method is illustrated in Fig.~\ref{fig2}.  

\subsection{Preliminaries}
A diffusion model is a form of generative model which has shown its impressive ability in a series of generative tasks~\cite{croitoru2023diffusion,cao2024survey}. It comprises a forward process that adds noise to a sample and a reverse process that removes noises~\cite{ho2020denoising}. Concretely, given the sample data ${\bm z}_{0}\in \mathbb{R}^{C\times H\times W}$ (where~$H$ and~$W$ are the height and width of the image spatial size,~$C$ denotes the channel size), the goal is to model the data distribution ${\bm z}_{0} \sim q\left ( {\bm z}_{0} \right )$ by a forward process, which iteratively adds Gussian noise into it like follows:
\begin{equation}
q\left ({\bm z}_{1:T} |{\bm z}_{0} \right ) := \prod_{t=1}^{T} q\left ({\bm z}_{t} |{\bm z}_{t-1} \right ), \label{1}
\end{equation}
\begin{equation}
q\left ({\bm z}_{t} |{\bm z}_{t-1} \right ) := N\left ({\bm z}_{t};\sqrt{1-\beta_{t}}{\bm z}_{t-1}, \beta_{t}{\bm I} \right ) \label{2}
\end{equation}
where ${\bm z}_{t}$ is the noise data at the timestep $t \in \left \{ 0,1,\cdot \cdot \cdot,T \right \} $, $\beta_{t} \in \left ( 0,1 \right ) $ defines a variance used at the timestep t. In addition, $\alpha_{t}:=1-\beta_{t}$ and $\bar{\alpha}_{t}:= {\textstyle \prod_{s=1}^{t}}\alpha_{s}$, which allow for Eq.~\eqref{1} and Eq.~\eqref{2} to be reformulated as: 
\begin{equation}
q({\bm z}_{t}|{\bm z}_{0}):=N({\bm z}_{t};\sqrt{\bar{\alpha}_{t}}{\bm z}_{0},(1-\bar{\alpha}_{t}){\bm I}) \label{3}
\end{equation}
therefore, the efficient sampling of ${\bm z}_{t}$ at arbitrary timestep $t$ in the Markov chain is expressed as:
\begin{equation}
{\bm z}_{t} =\sqrt{\bar{{\alpha}_{t}}} {\bm z}_{0} +\sqrt{1-\bar{\alpha}_{t}}\epsilon_{t} \label{4}
\end{equation}
where $\epsilon_{t} \in N(0,{\bm I})$. During training, our objective is to train a neural network $\Phi_{\theta}\left ({\bm z}_{t} ,t \right ) $, which predict the noise in
${\bm z}_{t}$ w.r.t. ${\bm z}_{0}$ by minimizing the L2 loss as follows:
\begin{equation}
L_{diff}=\left \| \epsilon_{t} -\Phi_{\theta }\left ({\bm z}_{t},t \right )  \right \|_{2}^{2} \label{5}
\end{equation}
during inference, the data sample ${\bm z}_{0}$ is reconstructed with an iterative denoising process using the trained network $\Phi_{\theta}$:
\begin{equation}
p\left ({\bm z}_{t-1} |{\bm z}_{t} \right ) := N\left ({\bm z}_{t-1};\Phi_{\theta}\left ({\bm z}_{t} ,t \right ), \sigma_{t}^{2}{\bm I} \right ) \label{6}
\end{equation}
where $\sigma_{t}^{2}$ represents the transition variance in DDIM~\cite{songdenoising}, it can accelerate the denosing process. In this paper, we leverage a diffusion model to eliminate the noises in low-light features with the help of normal-light features, which will be introduced in the next section.

\subsection{Diffusion-Guided Knowledge Distillation}
Low-light images exhibit significantly more incorrect predictions than normal-light images, demonstrating illumination's critical impact on segmentation. Their low contrast and blurred boundaries yield less distinct semantic information.
While Knowledge Distillation (KD) offers a potential solution by transferring semantic information from normal-light to low-light features, the inherent domain gap between these modalities limits the effectiveness of conventional KD approaches. Inspired by DiffKD~\cite{huang2024knowledge}, we conceptualize low-light features as noisy variants of their normal-light counterparts and employ a diffusion model to systematically reduce this noise. This enables low-light features to recover discriminative semantic information through denoising. Specifically, we propose training a diffusion model on normal-light features and the trained diffusion model is applied to noisy low-light features to generate denoised representations. Knowledge distillation is then applied between the denoised low-light features and normal-light ones for better feature alignment.

Formally, we use normal-light features ${\bm F}_{n}$ as teacher features in the forward noise process $q\left ({\bm F}_{n}^{t}\ |{\bm F}_{n} \right )$ (Eq.~\eqref{3}) to train the diffusion model with the loss function $L_{diff}$ (Eq.~\eqref{5}). Then dark features ${\bm F}_{d}$ are treated as student features and serve as the initial noisy input for the iterative denoising process of the trained diffusion model. Through this process, we obtain the denoised dark features $\bar{\bm F}_{d}$, which are then used to compute the KD loss with the normal-light features ${\bm F}_{n}$.
\begin{equation}
  L_{kd}=D(\bar{\bm F}_{d},{\bm F}_{n})
\end{equation}
where $D\left(\cdot \right)$ is a distance function. In our experiments, rather than relying on a single feature, we employ multiple features and final segmentation results to train the diffusion process. This hierarchical distillation approach, which transfers knowledge from shallow to deep features, enables more effective denoising of dark features and facilitates the transfer of richer semantic information to the low-light domain. 

\subsection{Depth-Guided Feature Fusion}
As illustrated in Fig.~\ref{fig2}, the details and structural information of objects in low-light images are often barely visible, significantly degrading their visual quality. To address this limitation and enrich the representation of dark features, we introduce depth maps ${\bm I}_{depth}$ of low-light images as additional visual priors. These depth maps, corresponding to the dark images, are generated using the Depth Anything model~\cite{yang2024depth}. As shown in Fig.~\ref{fig2}, the depth maps exhibit superior object details and boundary clarity compared to those from dark images, demonstrating their potential as valuable prior information to refine the distillation process. To effectively integrate these priors into the network,we employ Spatial Feature Transformation (SFT) layers~\cite{wang2018recovering} to encode the prior information as feature transformation parameters, which are then efficiently fused with the low-light features ${\bm F}_{d}$ to generate geometry-aware features ${\bm F}_{\bar d}$. The process can be formulated as follows:
\begin{equation}
prior=M\left ({\bm I}_{depth} \right )
\end{equation}
\begin{equation}
    \left ( \beta, \gamma  \right ) =\left (conv\left (conv\left (prior\right )\right ),conv\left (conv\left (prior\right )\right )\right )
\end{equation}
\begin{equation}
    {\bm F}_{\bar d} =SFT\left ( {\bm F}_{d}| \beta, \gamma\right )=\beta\odot {\bm F}_{d}+\gamma
\end{equation}
Here, $M$ means a series of convolutional layers with ReLU functions to extract the features as priors. $SFT(.)$ denotes Spatial Feature Transform layers, which utilize two convolutional layers respectively to learn a pair of parameters$(\beta, \gamma)$ and combine them with the dark features by scaling and shifting operations to get enhanced features ${\bm F}_{\bar d}$. Although the integration of depth priors improves performance, it inevitably introduces irrelevant background information, which may hinder the learning process. To address this issue and provide fine-grained depth guidance for learning dark features, we aim to preserve the potentially consistent feature regions with detailed information from both the original dark features ${\bm F}_{d}$ and the enhanced features ${\bm F}_{\bar d}$. Specifically, we employ sigmoid functions to highlight the activation regions of ${\bm F}_{\bar d}$ and ${\bm F}_{d}$, respectively. This allows us to learn an attention-guided map $A_{att}$, which captures consistent information from both ${\bm F}_{\bar d}$ and ${\bm F}_{d}$ while incorporating fine-grained details from the corresponding depth features. Finally, the enhanced features ${\bm F}_{fuse}$ are obtained by combining ${\bm F}_{\bar d}$ with the consistent activation regions derived from $A_{att}*({\bm F}_{d}+{\bm F}_{\bar d})$. This process can be formulated as follows:  
\begin{equation}
    att_{d}=\sigma \left ({\bm F}_{d}\right ),att_{\bar d}=\sigma \left ({\bm F}_{\bar d}\right )
\end{equation}
\begin{equation}
    A_{att}=\lambda \left (1-att_{d}\right )*\left (1-att_{\bar d}\right )+att_{d}*att_{\bar d}
\end{equation}
\begin{equation}
  {\bm F}_{fuse}={\bm F}_{\bar d}+A_{att}*\left ({\bm F}_{d}+{\bm F}_{\bar d}\right )  
\end{equation}
$\sigma\left (\cdot \right )$ means the Sigmoid function. $\lambda$ is a hyperparameter, which is set to $0.5$ in the experiment. To learn more comprehensive feature representations, we progressively integrate depth-based feature priors into the low-light backbone network layers. This forms a coarse-to-fine modulation chain that gradually refines feature expressions, learning fined-grained information while minimizing the introduction of irrelevant or noisy data.  

\subsection{Overall loss function}
The overall loss function is composed of the original classification and segmentation loss, diffusion losses and KD losses between normal-light features and low-light features. Noted that, the $L_{seg}$ here is a self-supervised segmentation loss. We employ the PAMR~\cite{araslanov2020single} to refine CAMs as pseudo-masks, which in turn supervise CAMs generation, forming a self-supervised segmentation loss.
\begin{equation}
    L_{overall}=L_{cls}+L_{seg}+\sum_{i=1}^{m}(L_{{diff}_{i}}+L_{{kd}_{i}})
\end{equation}
Here, $m$ is set to 3 in the experiment.
\section{Experiments}
\subsection{Datasets and Evaluation Metrics}\label{sec4:1}

\subparagraph{\textbf{Datasets.}}
We evaluate our method in the weakly supervised segmentation tasks based on synthetically darkened PASCAL VOC 2012~\cite{everingham2010pascal}, which is generated by the procedure of low-light synthesis pipeline~\cite{brooks2019unprocessing,cui2021multitask}. To further demonstrate the effectiveness of our method, we also conduct experiments on the realistically low-light LIS~\cite{chen2023instance} dataset. PASCAL VOC 2012 has~$21$ classes (including one background) of objects in total of~$4,369$ images, which are split of~$1,464$ images for training,~$1,449$ images for validation and~$1,456$ images for testing, respectively. Following the common practice in semantic segmentation, the augmented annotations from SBD~\cite{hariharan2011semantic} are used for an experimental comparison that has~$10,582$ training images. LIS dataset is consist of $9$ classes (including one background), which has $1561$ \emph{training} pairs with normal-light and low-light images and $669$ \emph{validation} pairs with normal-light and low-light images.

\begin{table}[t]
\begin{center}
\renewcommand\arraystretch{1.4}
\setlength{\tabcolsep}{.9pt}{
\caption{Segmentation performance on the \emph{val} set of synthetically darkened PASCAL VOC 2012 and \emph{test} set of realistically low-light LIS datasets. "tea." represents the results trained on normal-light images by the SSSS~\cite{araslanov2020single} model. "stu" represents the baseline results, which is directly trained on dark images by the SSSS~\cite{araslanov2020single} model. FLOPs is measured based on an input size $321\times 321$ during the inference stage.} 
\begin{tabular}{c|c|c|c|c|c}
\hline
\multirow{2}{*}{Datasets} & \multirow{2}{*}{\begin{tabular}[c]{@{}c@{}}Evaluation\\ Metrics\end{tabular}} & \multirow{2}{*}{tea.} & \multirow{2}{*}{stu.} & \multirow{2}{*}{+DGKD} & +DGKD \\
 &  &  &  &  & +DGF2 \\ \hline
\multirow{4}{*}{\begin{tabular}[c]{@{}c@{}}dark PASCAL \\ VOC 2012\end{tabular}} & mIoU(\%) & 59.7 & 43.4 & 55.2$_{\color{red}{+11.8}}$ & 57.1$_{\color{red}{+13.7}}$ \\
 & PixAcc(\%) & 88.4 & 81.1 & 87.4$_{\color{red}{+6.3}}$ & 87.9$_{\color{red}{+6.8}}$ \\ \cline{2-6} 
 & Params(M) & 138.0 & 138.0 & 148.0$_{\color{red}{+10.0}}$ & 148.3$_{\color{red}{+10.3}}$ \\
 & FLOPs(G) & 277.0 & 277.0 & 296.7$_{\color{red}{+19.7}}$ & 297.9$_{\color{red}{+20.9}}$ \\ \hline
\multirow{2}{*}{dark LIS} & mIoU(\%) & 57.7 & 43.9 & 52.2$_{\color{red}{+8.3}}$ & 54.1$_{\color{red}{+10.2}}$ \\
 & PixAcc(\%) & 87.4 & 78.1 & 87.2$_{\color{red}{+9.1}}$ & 86.5$_{\color{red}{+8.4}}$ \\ \hline
\end{tabular} \label{tab1}}
\vspace{-4mm}
\end{center}
\end{table}

\subparagraph{\textbf{Evaluation Metrics.}}
In the following experiments, we use the mean Intersection-over-Union (mIoU) as the evaluation metrics to evaluate the segmentation accuracy. Model parameters (Params) and FLOPs are also provided for evaluating the efficiency. 

\subsection{Experimental Settings.}
We adopt the single-stage segmentation framework SSSS~\cite{araslanov2020single} with a WideResNet38~\cite{wu2019wider} backbone as both the teacher model and student model. The teacher model is pre-trained on normal-light images. Since the LIS~\cite{chen2023instance} dataset has limited training samples, we augment it with PASCAL VOC 2012~\cite{everingham2010pascal} images corresponding to the $8$ shared categories of LIS. The teacher model is then trained on this augmented normal-light LIS dataset to ensure robust feature learning. We crop the image size to the $321\times 321$ and utilize the SGD optimizer with weight decay and momentum $5\times 10^{-4}$ and $0.9$, respectively. The initial learning rate is $0.005$ and the batch size is set to $6$. We distill knowledge from the normal-light features (Layer $3$ and Layer $6$) and the predicted segmentation maps of the teacher model to guide the learning of the low-light student model. 

\subsection{Ablation Study}
Our ablation experiments validate the effectiveness of the two proposed modules (i.e., DGKD and DGF2) on both the synthetically low-light PASCAL VOC 2012 and the real-world LIS datasets. 

\begin{table}[t] 
\begin{center}
\renewcommand\arraystretch{1.4}
\setlength{\tabcolsep}{.5pt}{
\caption{The results of ablation studies conducted on WSSS task to evaluate the segmentation performance on the \emph{val} set of sythetic PASCAL VOC 2012~\cite{everingham2010pascal}. ``mask" refers to the segmentation results generated by the model. }
\begin{tabular}{ l c  c c |  c c} 
Method &   &   &  & mIoU(\%) &PixAcc(\%) \\ 
\hline \hline
Baseline &    &    &     & 43.4 & 81.1 \\
\hline
\multicolumn{4}{l}{(\textbf{a}) \textbf{\emph{Superiority of DGKD}}} \\
\hline
+ MSE loss~\cite{paszke2019pytorch} &    &    &   &43.4$_{\color{red}{+0.0}}$ & 81.3$_{\color{red}{+0.2}}$\\
+ KL div loss~\cite{hinton2015distilling} &    &    &   &46.2$_{\color{red}{+2.8}}$ & 83.2$_{\color{red}{+2.1}}$\\
+ DIST loss~\cite{huang2022knowledge} &    &    &   &47.6$_{\color{red}{+4.2}}$ & 83.6$_{\color{red}{+2.5}}$\\
+ DGKD(features) &    &    &   &47.9$_{\color{red}{+4.5}}$ & 83.9$_{\color{red}{+2.8}}$\\
+ DGKD(mask) &    &    &   &53.8$_{\color{red}{+10.4}}$ & 87.0$_{\color{red}{+5.9}}$\\ 
+ DGKD(feature+mask) &    &    &   & 55.2$_{\color{red}{+11.8}}$ & 87.4$_{\color{red}{+6.3}}$ \\
\hline
\multicolumn{4}{l}{(\textbf{b}) \textbf{\emph{Superiority of DGF2}}} \\ 
\hline
+ DGKD + single SFT~\cite{wang2018recovering} &    &    &    & 56.1$_{\color{red}{+12.7}}$ & 87.8$_{\color{red}{+6.7}}$\\
+ DGKD + single DGF2 &    &    &    & 56.3$_{\color{red}{+12.9}}$ & 87.8$_{\color{red}{+6.7}}$\\
+ DGKD + multiple SFT~\cite{wang2018recovering} &    &    &    & 56.7$_{\color{red}{+13.3}}$ & 87.8$_{\color{red}{+6.7}}$\\
+ DGKD + multiple DGF2 &    &    &    & 57.1$_{\color{red}{+13.7}}$ & 87.9$_{\color{red}{+6.8}}$\\
\hline
\multicolumn{4}{l}{(\textbf{c}) \textbf{\emph{Superiority of Depth Anything}}} \\ 
\hline
+ DGKD + DGF2 with Depth Anything~\cite{yang2024depth} &    &    &    & 57.1$_{\color{red}{+13.7}}$ & 87.9$_{\color{red}{+6.8}}$\\
+ DGKD + DGF2 with ZoeDepth~\cite{bhat2023zoedepth} &    &    &    & 55.7$_{\color{red}{+12.3}}$ & 87.5$_{\color{red}{+6.4}}$\\
+ DGKD + DGF2 with MiDas~\cite{ranftl2020towards}  &    &    &    & 56.0$_{\color{red}{+12.6}}$ & 87.7$_{\color{red}{+6.6}}$\\
\hline
\end{tabular}\label{tab2}}
\vspace{-3mm}
\end{center}
\end{table}

The results in Table~\ref{tab1} demonstrate the effectiveness of the proposed DGKD and DGF2 modules. We can observe that, on the dark PASCAL VOC 2012 dataset, the teacher network, trained on well-illuminated images using the SSSS~\cite{araslanov2020single} model, achieves an mIoU of $59.7\%$ and a PixAcc of $88.4\%$. While the baseline student network, without any additional modules, achieves an mIoU of $43.4\%$ and a PixAcc of $81.1\%$. The significant performance gap compared to the teacher model highlights the challenge of weakly supervised segmentation in low-light conditions. When the DGKD module is added to the student network, the mIoU improves by $11.8\%$, and the PixAcc increases to $87.4\%$. This confirms DGKD’s ability to recover semantic information obscured by illumination noise and align low-light and normal-light features effectively. When both DGKD and DGF2 modules are incorporated, the student network achieves the highest performance, with an mIoU of $57.1\%$, nearly closing the gap to normal-light performance. The results clearly show that both DGKD and DGF2 modules contribute significantly to improving segmentation performance on low-light images. The DGKD module alone addresses semantic confusion caused by low light. And the DGF2 module complements DGKD by enhancing structural feature learning. Similar significant improvements are observed on the real-world LIS~\cite{chen2023instance} dataset, demonstrating the generalized ability of our method to realistically low-light scenarios.
To further demonstrate the superiority of DGKD and DGF2 modules, we compare them against other refinement strategies.The experimental results are summarized in Table~\ref{tab2}.
\begin{figure}[tb]
\includegraphics[width=.48 \textwidth]{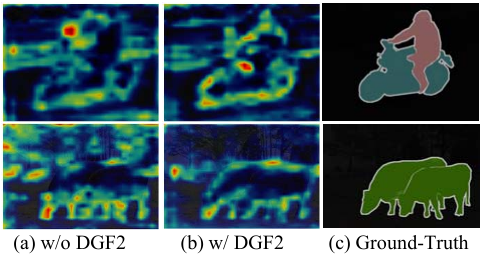}
\vspace{-6mm}
\caption{The feature visualization comparison (with vs. without DGF2) shows that DGF2 brings structural learning improvements to the DGKD-WLSS model.}
\label{fig3}
\vspace{-5mm}
\end{figure}
\subparagraph{\textbf{Superiority of DGKD.}} Our DGKD module utilizes multiple intermediate features and output mask to conduct distillation. To validate the superiority of DGKD, we compare it with different KD losses~\cite{paszke2019pytorch,huang2022knowledge,hinton2015distilling} without denoising operation. When using only intermediate features with DIST distillation loss~\cite{huang2022knowledge}, the model achieves $47.6\%$ mIoU and $83.6\%$ PixAcc, showing that conventional KD improves performance, but remains limited by noisy low-light features. The subsequent application of denoising to these intermediate features can improve by $0.3\%$ in both mIoU and PixAcc. In particular, when we focus solely on distilling and denoising the segmentation masks, performance improves substantially to $53.8\%$ mIoU and $87.0\%$ PixAcc, highlighting the importance of semantic-level alignment through denoising. The best results emerge when combining denoising and distillation for both intermediate features and segmentation masks, achieving $55.2\%$ mIoU and $87.4\%$ PixAcc and outperforming the baseline by $13.1\%$ mIoU while adding only $10.3$M parameters (shown in Table~\ref{tab1}). These results show that our approach of denoising and distilling both hierarchical features and final predictions enables superior feature alignment between low-light and normal-light domains, leading to substantially improved segmentation performance. The controlled parameter growth makes this performance gain particularly efficient, validating the practical value of our DGKD design.
\begin{table}[t!]
\begin{center}
\renewcommand\arraystretch{1.4}
\setlength{\tabcolsep}{6pt}{
\caption{Comparisons on the \emph{val} set of synthetic low-light PASCAL VOC 2012 dataset~\cite{everingham2010pascal} with the state-of-the-art methods originally proposed for normal-light images. These methods were implemented without any modifications and were retrained on our synthetic low-light dataset to ensure fair comparison. ``Seg. Backbone'' denotes the backbone network used for the semantic segmentation task. }
\begin{tabular}{ l c | c c } 
Methods & Seg. Backbone & \emph{mIoU}~(\%) & \emph{PixAcc}~(\%) \\
\hline \hline
SSSS~\cite{araslanov2020single} & WideResNet38  &30.8  & 55.4 \\
WS-FCN~\cite{wang2023coupling} & WideResNet38   & 37.6  & 73.1 \\
AFA~\cite{ru2022learning} & MiT-B1   & 47.0  & 79.7 \\
SLRNet~\cite{pan2022learning} & WideResNet38  & 45.8  & 83.5 \\
ToCo~\cite{ru2023token} &  ViT-B   & 39.3  & 79.0 \\
WeCLIP~\cite{zhang2024frozen} &  ViT-B   & 32.3  & 78.7 \\
\cdashline{1-4}[0.8pt/2pt]
\textbf{DGKD-WLSS} & WideResNet38 & \textbf{57.1}   & \textbf{87.9} \\
\hline 
\end{tabular}
\label{tab3}}
\vspace{-3mm}
\end{center}
\end{table}


\subparagraph{\textbf{Superiority of DGF2.}} As shown in Table~\ref{tab2}, the introduction of a single SFT~\cite{wang2018recovering} module can achieve $56.1\%$ mIoU and $87.8\%$ PixAcc, confirming that integrating the depth maps into the student network is helpful. Single DGF2 further refines this with $56.3\%$ mIoU, indicating that our proposed DGF2 better leverages depth priors for feature enhancement. Multiple DGF2 achieve the best performance of $57.1\%$ mIoU and $87.9\%$ PixAcc, proving that hierarchical depth guidance is essential for structured feature learning. Compared with performance of the DGKD module, DGF2 can further improve by $1.9\%$ mIoU and $0.5\%$ PixAcc with the only addition of $0.3$ M parameters. Besides, Fig.~\ref{fig3} intuitively shows the performance improvements in structural learning. This confirms that depth guidance complements semantic distillation, particularly in recovering fine-grained structures. These results demonstrate that explicit geometric guidance is important under illumination degradation.
In addition, we utilize the depth maps generated by the Depth Anything~\cite{yang2024depth} model because it has stronger generalization capabilities and can generate relatively accurate depth maps even for unseen images. In order to show the robustness of DGF2, we conducted experiments using depth maps generated by Depth Anything~\cite{yang2024depth} and two other weaker depth estimation methods, ZoeDepth~\cite{bhat2023zoedepth} and MiDaS~\cite{ranftl2020towards} shown in Table~\ref{tab2}. We can see that all depth maps generated by these methods can help to improve the segmentation performance. More accurate depth maps can help produce better segmentation performance, and inaccurate or noisy depth maps may not provide significant benefits.

\begin{table}[t]
\begin{center}
\renewcommand\arraystretch{1.4}
\setlength{\tabcolsep}{5pt}{
\caption{Quantitative comparisons of our method with the other enhancement methods. To show the generalization, the SSSS~\cite{araslanov2020single} model are pre-trained by only the normal \emph{train} set of PASCAL VOC 2012~\cite{everingham2010pascal} and evaluated on \emph{test} set of the LIS~\cite{chen2023instance} dataset. Our method is trained on sythetically dark \emph{train} set of PASCAL VOC 2012~\cite{everingham2010pascal} and its performance is directly evaluated on the dark \emph{test} set of LIS~\cite{chen2023instance}. }
\begin{tabular}{ l c |c c } 
Methods & Seg. Method & mIoU(\%)  & PixAcc (\%)\\ 
\hline \hline
\multicolumn{4}{l}{(\textbf{a}) \textbf{\emph{Direct}}} \\
\cdashline{1-4}[0.8pt/2pt]
-- & SSSS & 34.5 & 83.0 \\
\hline 
\multicolumn{4}{l}{(\textbf{b}) \textbf{\emph{Enhance}}} \\ 
\cdashline{1-4}[0.8pt/2pt]
HE~\cite{gonzales1987digital}& SSSS & 34.7 & 82.9\\
Retinex-Net~\cite{wei2018deep} & SSSS & 30.8 & 82.3 \\
EnlightenGAN~\cite{jiang2021enlightengan} & SSSS & 38.9 & 83.8 \\
Zero-DCE~\cite{guo2020zero} & SSSS & 40.2 & 84.1 \\ 
WCDM~\cite{jiang2023low} & SSSS & 39.6 & 84.1 \\ 
HVI~\cite{yan2025hvi}& SSSS & 37.1 & 83.4 \\
\hline 
\multicolumn{4}{l}{(\textbf{c}) \textbf{\emph{Integrated enhance}}} \\ 
\cdashline{1-4}[0.8pt/2pt]
CNNPP~\cite{liu2023improving} & SSSS &38.0  & 73.5 \\
\hline 
\multicolumn{4}{l}{(\textbf{d}) \textbf{\emph{Distillation}}} \\ 
\cdashline{1-4}[0.8pt/2pt]
\textbf{DGKD-WLSS} (ours) & SSSS  & \textbf{46.3} & \textbf{84.1}  \\
\hline
\end{tabular}
\label{tab4}}
\end{center}
\vspace{-6mm}
\end{table}


\begin{figure*}[tb]
\includegraphics[width=.98 \textwidth]{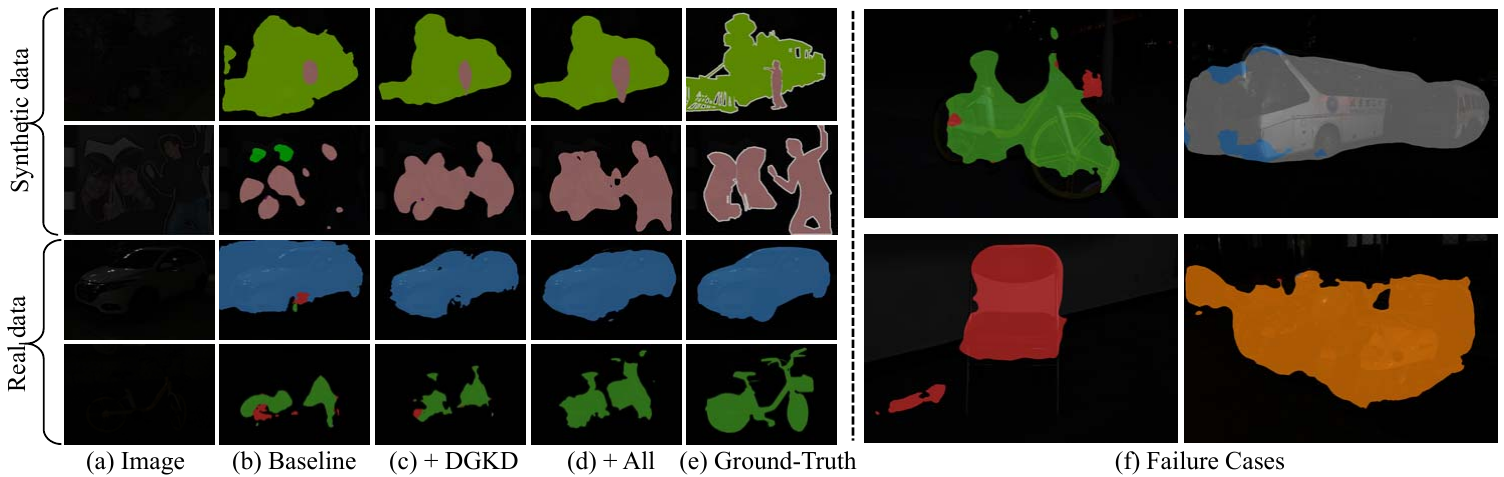} 
\vspace{-2mm}
\caption{Visualizations of segmentation masks generated by various methods, including the baseline, +DGKD, +All (\ie, + DGKD+DGF2) and ground-truth (left). The top two rows of the left display the segmentation results on the \emph{val} set of synthetically darkened PASCAL VOC 2012~\cite{everingham2010pascal}. The last two rows show the results on the \emph{test} set of realistically dark LIS dataset. These results demonstrate the effectiveness of our proposed modules in progressively improving the representation of low-light features. At the same time, we exhibit the failure cases for visualization (right). }
\label{fig4}
\vspace{-3mm}
\end{figure*}

\subparagraph{\textbf{Quantitative Results.}} 
The comparative results in Table~\ref{tab3} demonstrate the significant challenges faced by state-of-the-art weakly supervised semantic segmentation methods when applied to low-light conditions. All methods, originally designed for normal-light images and implemented without modifications, exhibited substantial performance degradation when retrained on our synthetically darkened PASCAL VOC 2012 dataset. The conventional SSSS~\cite{araslanov2020single} method, while effective in normal lighting 
, only reached $30.8\%$ mIoU in low-light conditions, revealing the severe impact of illumination degradation on segmentation performance. More recent approaches like AFA~\cite{ru2022learning} and SLRNet~\cite{pan2022learning} showed improved but still limited results, indicating that current architectures lack effective mechanisms to handle low-light challenges. Our proposed DGKD-WLSS method significantly outperformed all methods, achieving $57.1\%$ mIoU and $87.9\%$ PixAcc. It improves by $10.1\%$ mIoU and $8.2\%$ PixAcc compared with the best performing AFA method. This substantial performance gap highlights the effectiveness of our novel components: the diffusion-guided knowledge distillation for semantically consistent feature learning and depth-aware feature fusion for structural preservation. The results clearly demonstrate that simply adapting normal-light methods to low-light conditions through retraining is insufficient, and that specialized approaches addressing illumination-specific challenges are crucial for achieving robust performance in low-light semantic segmentation tasks. Notably, DGKD-WLSS's strong performance was achieved using the same WideResNet-38 backbone as several methods like WS-FCN~\cite{wang2023coupling}, SLRNet~\cite{pan2022learning}, confirming that our architectural innovations rather than backbone capacity are responsible for the improvements.

For a comprehensive quantitative comparison with other enhancement methods, we employed the weakly supervised single-stage SSSS approach~\cite{araslanov2020single} as our unified segmentation framework. To rigorously evaluate the practical effectiveness and generalized ability of our method in real-world low-light conditions, all comparative methods were trained on the training set of PASCAL VOC 2012~\cite{everingham2010pascal}, which shares the same $8$ categories as the LIS dataset, and tested on the test set of LIS dataset. Our method was trained on synthetically darkened PASCAL VOC 2012 and evaluated on the realistically low-light LIS dataset. While competing enhancement methods first enhanced the dark LIS images and then fed the improved results to the SSSS model trained on the original PASCAL VOC 2012 data for evaluation. This standardized evaluation protocol ensures fair and consistent comparison of different approaches under low-light conditions. The experimental results, summarized in Table~\ref{tab4}, demonstrate the superior performance of our method on the real LIS dataset. The baseline SSSS model without any enhancement achieved only $34.5\%$ mIoU and $83.0\%$ PixAcc on the low-light LIS test set, highlighting the inherent challenges of segmenting unprocessed low-light images. Traditional histogram equalization (HE)~\cite{gonzales1987digital} provided minimal improvement, indicating the limited utility of basic contrast enhancement for segmentation tasks. Among separately trained enhancement methods, Zero-DCE~\cite{guo2020zero} performed best with $40.2\%$ mIoU and $84.1\%$ PixAcc, demonstrating its strong low-light enhancement capabilities. The jointly trained CNNPP~\cite{liu2023improving} method achieved $38.0\%$ mIoU but suffered reduced PixAcc, suggesting potential artifact introduction despite its semantic improvement. Our proposed DGKD-WLSS method outperformed all methods, achieving state-of-the-art results of $46.3\%$ mIoU and $84.1\%$ PixAcc. This significant performance gain validates the effectiveness of our DGKD-WLSS.
The consistent improvements across both synthetic and real-world low-light scenarios further demonstrate the robustness and generalized ability of our framework.

\subparagraph{\textbf{Qualitative Results.}} 
Fig.~\ref{fig4} presents an intuitively visual comparison between the baseline and our proposed DGKD-WLSS on both synthetic and real-world datasets, clearly demonstrating the progressive improvement contributed by each module. We can observe that DGKD module substantially enhances segmentation accuracy compared to the baseline, particularly in recovering semantically meaningful regions that were previously obscured by low-light noise. The integration of DGKD and DGF2 produces segmentation masks that progressively approximate the ground truth annotations, with particularly notable improvements in challenging areas involving fine structures and low-contrast boundaries. These visual comparisons provide compelling evidence for the effectiveness of our approach in addressing the challenges of low-light semantic segmentation, including semantic ambiguity and structural feature degradation.

Besides, our method still suffers from segmentation deficiencies shown on the right side of Figure~\ref{fig4}, which can be categorized into (1) misclassifying detailed parts of objects as other categories (e.g., “bus” windows and undercarriages being labeled as “cars”) and (2) incomplete segmentation of fine-grained object structures (e.g., “chair” legs). These issues stem from two primary causes: First, in low-light conditions, dark regions lose critical details or become blurred, triggering false activations that lead to misclassification. While DGKD-WLSS’s diffusion-guided denoising knowledge distillation restores most semantic knowledge, it struggles with subtle features. Second, noise corruption under dim lighting obscures discriminative textures between similar objects, hampering model judgment. A potential solution involves integrating image enhancement techniques to improve initial visual representation before applying DGKD-WLSS for feature-level refinement. 

\section{Conclusion}
In this paper, we propose a novel approach, DGKD-WLSS, for weakly-supervised low-light semantic segmentation. Our method transfers semantic knowledge from normal-light images to low-light images by treating low-light (student) features as noisy versions of normal-light (teacher) features. From this perspective, we leverage a diffusion model to effectively denoise the low-light features, which improves the quality of segmentation masks for low-light images. Furthermore, we introduce depth maps as additional visual priors 
to provide the structural information, enabling better feature representation. 
Extensive experiments demonstrate the effectiveness of our proposed method. In the future work, we plan to explore low-light semantic segmentation in scenarios where paired normal-light and low-light images are unavailable, aiming to further extend the applicability and robustness of our approach.

\newpage
\section{Acknowledgments}
This work was supported  by the National
Natural Science Foundation of China under Grants 62332010.
\bibliographystyle{ACM-Reference-Format}
\balance
\bibliography{main}
\newpage
\appendix

\section{Supplementary Materials }
\subsection{Results of backbone with ViT }
To ensure a fair comparison with methods using WideResNet38 as the backbone (e.g., SSSS, WS-FCN, and SLRNet in Table~\ref{tab3} of the paper), we also adopt WideResNet38 as our backbone network. In addition, we can see that although we take the WideResNet38 as the backbone, our method can perform better than methods using MiT-B,ViT-B as backbones such as AFA, ToCo. It validates the effectiveness of our model design instead of depending on the stronger backbone. To valid the effectiveness of our method, we conduct an experiment with ViT-B as the backbone, which can obtain 61.9\% mIoU and 89.9\% PixAcc on the val set of synthetic PASCAL VOC 2012 as shown in Table~\ref{A1}. It demonstrates that stronger backbone can further improve the segmentation performance of DGKD-WLSS.
\begin{table}[htb]
\begin{center}
\renewcommand\arraystretch{1.4}
\setlength{\tabcolsep}{5pt}{
\vspace{-3mm}
\caption{Segmentation performance of DGKD-WLSS with different backbones.}
\begin{tabular}{l|l|l}
Settings & mIoU (\%) & PixAcc (\%) \\
\hline \hline
DGKD-WLSS with WideResNet38 & 57.1 & 87.9 \\
\hline
DGKD-WLSS with ViT-B & 61.9 & 89.9 \\
\hline
\end{tabular}
\label{A1}}
\end{center}
\vspace{-3mm}
\end{table}

\subsection{Computational overhead of diffusion model}
During training, introducing diffusion models into the knowledge distillation framework incurs 124.7G FLOPs when the input size is 321×321. Concretely, we perform one (T=1) forward pass of teacher features (or pseudo-masks) to train the noise prediction network, followed by five(T=5) forward passes to denoise student features. These six times of forwarding bring computational overhead. However, the diffusion model is not utilized during model inference, thus preserving deployment efficiency.

\subsection{Results of a two-stage experiment}
we add a two-stage experiment and utilize the pseudo-masks of DGKD-WLSS model to train the segmentation model (deeplabv2 with resnet101 backbone). In addition, to quantitatively evaluate the effectiveness of our method, we also compared it with two recent state-of-the-art multi-stage WSSS approaches, CTI~\cite{yoon2024class} and S2C~\cite{kweon2024sam}. As shown in Table~\ref{A2}, we can see that our two-stage segmentation results can obtain 58.2\% mIoU and 88.2\% PixAcc, which perform better than the single-stage results. In addition, two-stage weakly supervised semantic segmentation methods,  CTI~\cite{yoon2024class} and S2C~\cite{kweon2024sam}, originally designed for normal-light images and implemented without modifications, exhibit substantial performance degradation when retraining on our synthetically darkened PASCAL VOC 2012 dataset. These results validate that severe impact of illumination degradation on segmentation performance while our method can alleviate this problem a lot.

\subsection{Train on Cityscapes and evaluate on NightCity}
we supplement the fully supervised semantic segmentation experiments, where models trained on synthetically darkened Cityscapes~\cite{cordts2016cityscapes} are evaluated on NightCity~\cite{tan2021night} in Table~\ref{A3}. In addition, we also provide the segmentation results of training on normal Cityscapes and testing on NightCity. The results demonstrate that training on synthetically darkened Cityscapes yields better segmentation performance on NightCity than directly applying models trained on normal-light Cityscapes. This validates our method's effectiveness for low-light image segmentation.
\begin{table}[htb]
\begin{center}
\renewcommand\arraystretch{1.4}
\setlength{\tabcolsep}{5pt}{
\caption{ Comparisons on the val set of synthetic low-light PASCAL VOC 2012 dataset with the state-of-the-art multi-stage WSSS methods originally proposed for normal-light images.}
\begin{tabular}{l|l|l}
Methods & mIoU (\%) & PixAcc (\%) \\
\hline \hline
DGKD-WLSS(single-stage) & 57.1 & 87.9 \\
\hline
DGKD-WLSS(two-stage) & 58.2 & 88.2 \\
\hline
CTI~\cite{yoon2024class} & 28.4 & 69.8 \\
\hline
S2C~\cite{kweon2024sam} & 46.4 & 83.2 \\
\hline
\end{tabular}
\label{A2}}
\end{center}
\vspace{-3mm}
\end{table}
\begin{table}[htb]
\begin{center}
\renewcommand\arraystretch{1.4}
\setlength{\tabcolsep}{5pt}{
\caption{Performance of training on (synthetically dark) Cityscapes~\cite{cordts2016cityscapes} and testing on low-light NightCity~\cite{tan2021night}.}
\begin{tabular}{l|l|l}
Settings & mIoU (\%) & PixAcc (\%) \\
\hline \hline
Train on Cityscapes & 18.8 & 56.1 \\
\hline
Train on synthetically dark Cityscapes & 22.7 & 62.1 \\
\hline
\end{tabular}
\label{A3}}
\end{center}
\vspace{-3mm}
\end{table}

\subsection{Impact of the diffusion timestep T and the depth fusion hyperparameter \texorpdfstring{$\lambda$}{lambda}}
We perform an ablation study to evaluate the impact of parameters T and $\lambda$, with the experimental results presented in Table~\ref{A4} and Table~\ref{A5}, respectively. As shown in the tables, the optimal segmentation performance is achieved when $\lambda = 0.5$ and $T = 5$, demonstrating the effectiveness of these parameter settings.
\begin{table}[htb]
\begin{center}
\renewcommand\arraystretch{1.4}
\setlength{\tabcolsep}{5pt}{
\caption{Ablation study of different \texorpdfstring{$\lambda$}{lambda}.}
\begin{tabular}{l|l|l|l}
\hline
$\lambda$ & 0.4 & 0.5 & 0.6 \\
\hline
mIoU (\%) & 56.6 & 57.1 & 56.1 \\
\hline
PixAcc (\%) & 87.9 & 87.9 & 87.8 \\
\hline
\end{tabular}
\label{A4}}
\end{center}
\vspace{-3mm}
\end{table}
\begin{table}[htb]
\begin{center}
\renewcommand\arraystretch{1.4}
\setlength{\tabcolsep}{5pt}{
\caption{Ablation study of different $T$.}
\begin{tabular}{l|l|l|l}
\hline
T & 4 & 5 & 6 \\
\hline
mIoU (\%) & 57.0 & 57.1 & 57.1 \\
\hline
PixAcc (\%) & 87.9 & 87.9 & 87.9 \\
\hline
\end{tabular}
\label{A5}}
\end{center}
\vspace{-3mm}
\end{table}

\end{document}